\documentclass[letterpaper]{IEEEtran}
\usepackage[latin9]{inputenc}
\usepackage{color}
\usepackage{booktabs}
\usepackage{url}
\usepackage{amsmath}
\usepackage{amssymb}
\usepackage{graphicx}
\usepackage{hyperref}
\makeatletter

\pdfpageheight\paperheight
\pdfpagewidth\paperwidth

\providecommand{\tabularnewline}{\\}

\usepackage{cvpr}
\usepackage{times}
\usepackage{epsfig}
\usepackage{graphicx}

\usepackage{authblk}
\author[1]{Amir Rosenfeld}
\author[2]{Richard Zemel}
\author[1]{John K. Tsotsos}

\affil[1]{York University}
\affil[2]{University of Toronto}
\affil[1,2]{Toronto, Canada}
\date{}                     %% if you don't need date to appear
 \cvprfinalcopy % *** Uncomment this line for the final submission

\def\cvprPaperID{****} % *** Enter the CVPR Paper ID here

\@ifundefined{showcaptionsetup}{}{%
 \PassOptionsToPackage{caption=false}{subfig}}
\usepackage{subfig}
\makeatother

\begin{document}

\title{The Elephant in the Room}
\maketitle
\begin{abstract}
We showcase a family of common failures of state-of-the art object
detectors. These are obtained by replacing image sub-regions by another
sub-image that contains a trained object. We call this ``object
transplanting''. Modifying an image in this manner is shown to have
a non-local impact on object detection. Slight changes in object position
can affect its identity according to an object detector as well as
that of other objects in the image. We provide some analysis and suggest
possible reasons for the reported phenomena.
\end{abstract}

\section*{Introduction}

Reliable systems for image understanding are crucial for applications
such as autonomous driving, medical imaging, etc.

Adversarial examples \cite{szegedy2013intriguing} have been suggested
as small targeted perturbations. We show another kind of perturbation.
As opposed to adversarial examples, these are not nor	m-bounded. They
involve putting (``transplanting'') an object from one image in
a new location of another image. This is shown to have multiple effects
on the object detector. We demonstrate this phenomena through a series
of experiments, suggesting some possible explanations. 

\section*{Experiments}

We begin with some qualitative results. Figure \ref{fig:Detecting-an-elephant}
(a) shows the results of a state-of-the-art object detection method
(Faster-RCNN \cite{ren2015faster} with a NASNet backbone \cite{zoph2017learning})
when applied to an image of a living-room from the Microsoft COCO
object detection benchmark \cite{lin2014microsoft} on which the detector
was trained. Using the ground-truth we extract an object (elephant)
along with its mask from another image and ``transplant'' it into
this image of a living-room at various locations. We refer to the
transplanted object as $T$. The results can be seen in sub-figures
\emph{\ref{fig:Detecting-an-elephant} b-l. }We note several interesting
phenomena as the object $T$ translates along the image:
\begin{enumerate}
\item Detection is not stable: the object may occasionaly become undetected
or be detected with sharp changes in confidence
\item The reported identity of the object $T$ is not consistent (chair
in \ref{fig:Detecting-an-elephant},\emph{f}): the object may be detected
as a variety of different classes depending on location
\item The object causes non-local effects: objects non-overlapping with
$T$ can switch identity, bounding-box, or disappear altogether.
\end{enumerate}

\noindent \begin{center}
\begin{figure*}
\noindent \begin{centering}
\subfloat[]{\begin{centering}
\includegraphics[viewport=0bp 100bp 300bp 400bp,clip,width=0.33\textwidth]{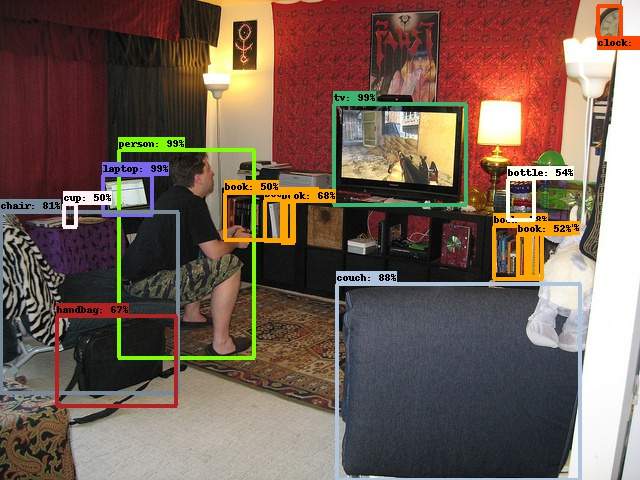}
\par\end{centering}
}\subfloat[]{\begin{centering}
\includegraphics[viewport=0bp 100bp 300bp 400bp,clip,width=0.33\textwidth]{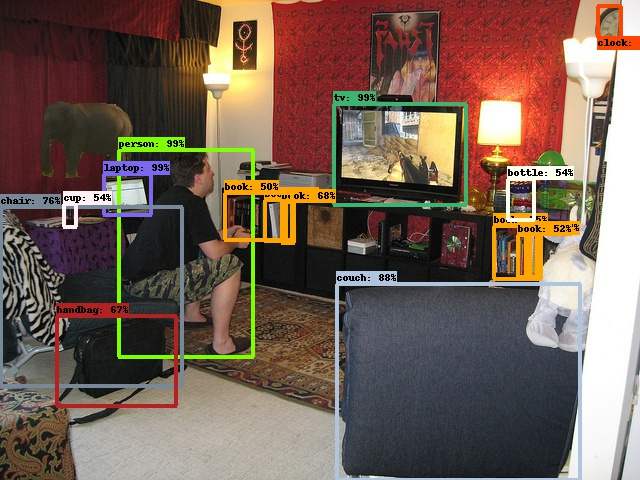}
\par\end{centering}
}\subfloat[]{\begin{centering}
\includegraphics[viewport=0bp 100bp 300bp 400bp,clip,width=0.33\textwidth]{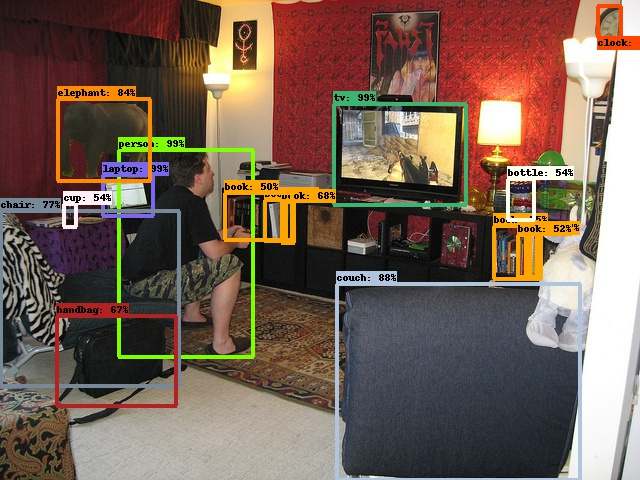}
\par\end{centering}
}
\par\end{centering}
\noindent \begin{centering}
\subfloat[]{\begin{centering}
\includegraphics[viewport=0bp 100bp 300bp 400bp,clip,width=0.33\textwidth]{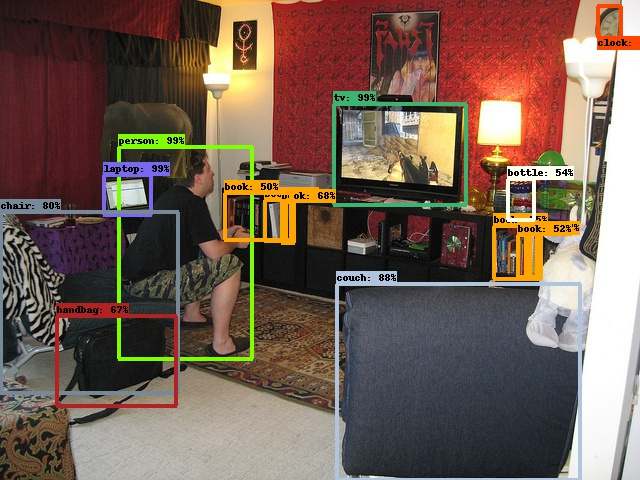}
\par\end{centering}
}\subfloat[]{\begin{centering}
\includegraphics[viewport=0bp 100bp 300bp 400bp,clip,width=0.33\textwidth]{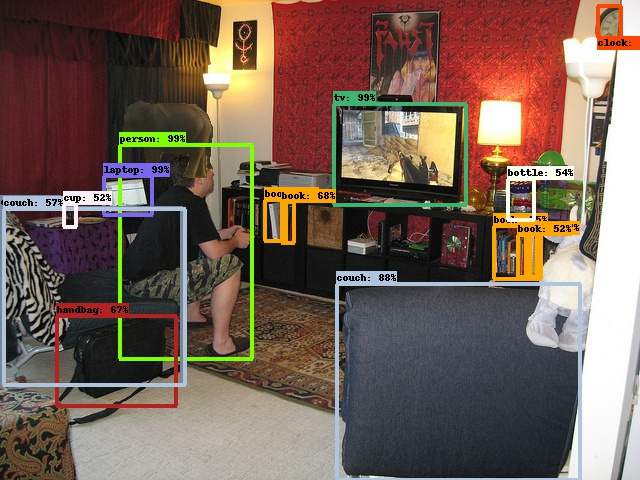}
\par\end{centering}
}\subfloat[]{\begin{centering}
\includegraphics[viewport=0bp 100bp 300bp 400bp,clip,width=0.33\textwidth]{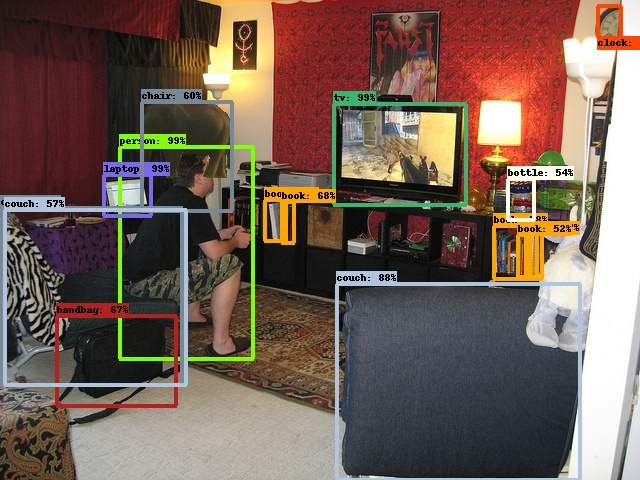}
\par\end{centering}
}
\par\end{centering}
\noindent \begin{centering}
\subfloat[]{\begin{centering}
\includegraphics[viewport=0bp 100bp 300bp 400bp,clip,width=0.33\textwidth]{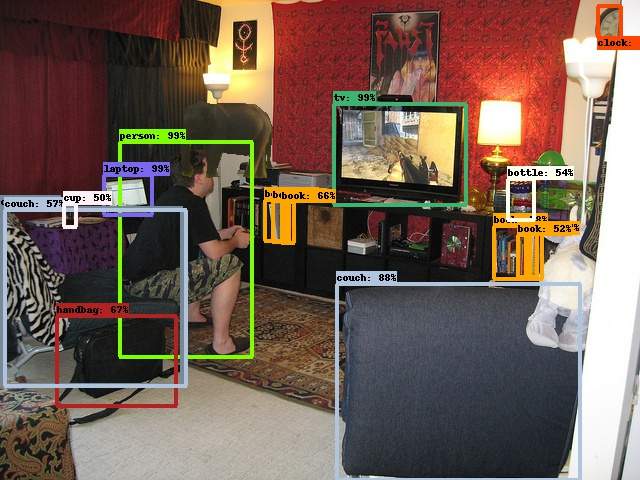}
\par\end{centering}
}\subfloat[]{\begin{centering}
\includegraphics[viewport=0bp 100bp 300bp 400bp,clip,width=0.33\textwidth]{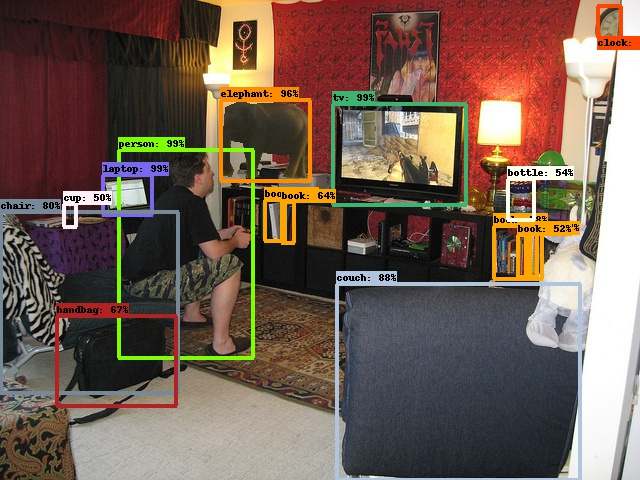}
\par\end{centering}
}\subfloat[]{\begin{centering}
\includegraphics[viewport=0bp 100bp 300bp 400bp,clip,width=0.33\textwidth]{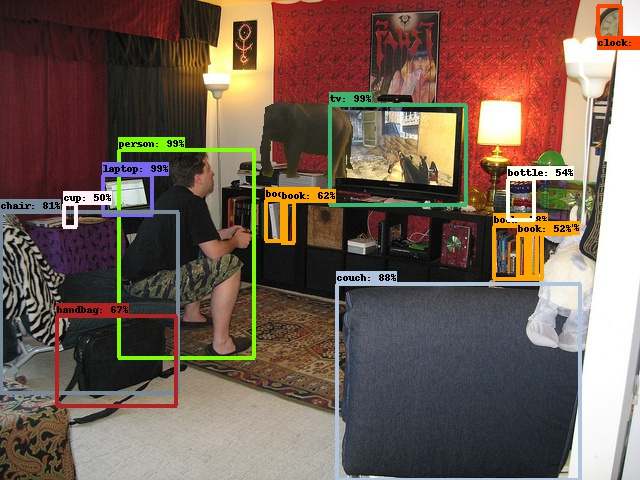}
\par\end{centering}
}
\par\end{centering}
\centering{}\caption{\label{fig:Detecting-an-elephant}Detecting an elephant in a room.
A state-of-the-art object detector detects multiple images in a living-room
(\emph{a}). A transplanted object (elephant) can remain undetected
in many situations and arbitrary locations (\emph{b,d,e,g,i}). It
can assume incorrect identities such as a chair (\emph{f}). The object
has a non-local effect, causing other objects to disappear (cup, \emph{d,f,
}book, \emph{e-i} ) or switch identity (chair switches to couch in
\emph{e}). It is recommended to view this image in color online.}
\end{figure*}
\par\end{center}

\subsection*{Detailed Analysis}

We now present detailed experiments to further demonstrate each of
these phenomena. All of our experiments use images taken from the
validation set of the 2017 version of the MS-COCO dataset. Unless
otherwise specified, we use models from the Tensorflow Object Detection
API \cite{huang2017speed}. This enables easy reproducibility of our
experiments and access to a diverse set of contemporary state-of-the-art
object detection architectures. Unless specified otherwise, we only
use models that were trained on MS-COCO. The models are downloaded
from the corresponding API's webpage\footnote{\url{https://github.com/tensorflow/models/blob/master/research/object_detection/g3doc/detection_model_zoo.md}}
and applied to images using the officially provided code. Table \ref{tab:Models-used-in}
specifies the models we used. 

\paragraph{Test Image Generation }

As the example in Fig. \ref{fig:Detecting-an-elephant} may seem a
bit contrived, we provide further examples which are generated randomly.
In short, each example is created by picking a random pair of images
$I,J$ and transplanting a random object from the image $J$ into
image $I,$then testing the effect on object detection. This is done
as described below. 

A test image $I_{t}^{x,y}$ is generated as follows: we pick an image
$J\neq I.$ Using the ground-truth, we randomly select one of the
object instances $Obj_{i}$ along with its provided segmentation mask
$M_{i}$. We ``transplant'' $Obj_{i}$ to a location whose origin
is $t_{x},t_{y}$ in image $I$. Denote by $T_{x,y}$ the translated
object (abbreviated as $T$ where specifying $x,y$ is unnecessary).
We copy each foreground pixel using the segmentation masks $M_{i}$.
All other pixels are unmodified. The $t_{x},t_{y}$ translations are
varied so that $t_{x}\in[0,k,k+1...,w_{0}]$ and $t_{y}\in[0,k,k+1...,h_{0}]$,
where $k$ is a step-size in pixels ($k=10$) and $w_{o},h_{0}$ are
such that the transplanted object is always fully inside the test
image $I_{t}^{x,y}$. For typical images, whose side is hundreds of
pixels, this creates about a thousand different test images. 

The image pairs $I,J$ are all picked randomly from the validation
set. We discard pairs for which the transplanted object is too large
(allowing a degenerate amount of translations) or too small, without
any distinguishing visual features except its context. 

The detector's output is a set of detections $\mathcal{D}_{x,y}$=$\{d_{m}\},m\in1\dots M$
for each test image $I_{t}^{x,y}$. Each detection $d_{m}=<b,s,c>_{m}$
is a made of bounding box coordinates $b$, detection score $s$ and
object category $c$. $M$ is the number of detections for the image. 

We denote by $\mathcal{D_{\emptyset}}$ the detections on the original
image, where no object was transplanted. 

\paragraph*{Matching Detections}

We analyze the effect of the transplanted object on $I$ by comparing
$\mathcal{D}_{x,y}$ to $\mathcal{D_{\emptyset}}$. One simple method
of doing so is to compare the set of confidently detected object classes.
Let $C_{x,y}$ and $C_{\ensuremath{\emptyset}}$ be the corresponding
sets of detected object categories. The cardinality of the difference
$\left|C_{x,y}\backslash C_{\ensuremath{\emptyset}}\right|$ is used
to sort the detection results. We call this the ``Class-Matching''
criteria. 

\begin{figure*}
\includegraphics[width=1\textwidth]{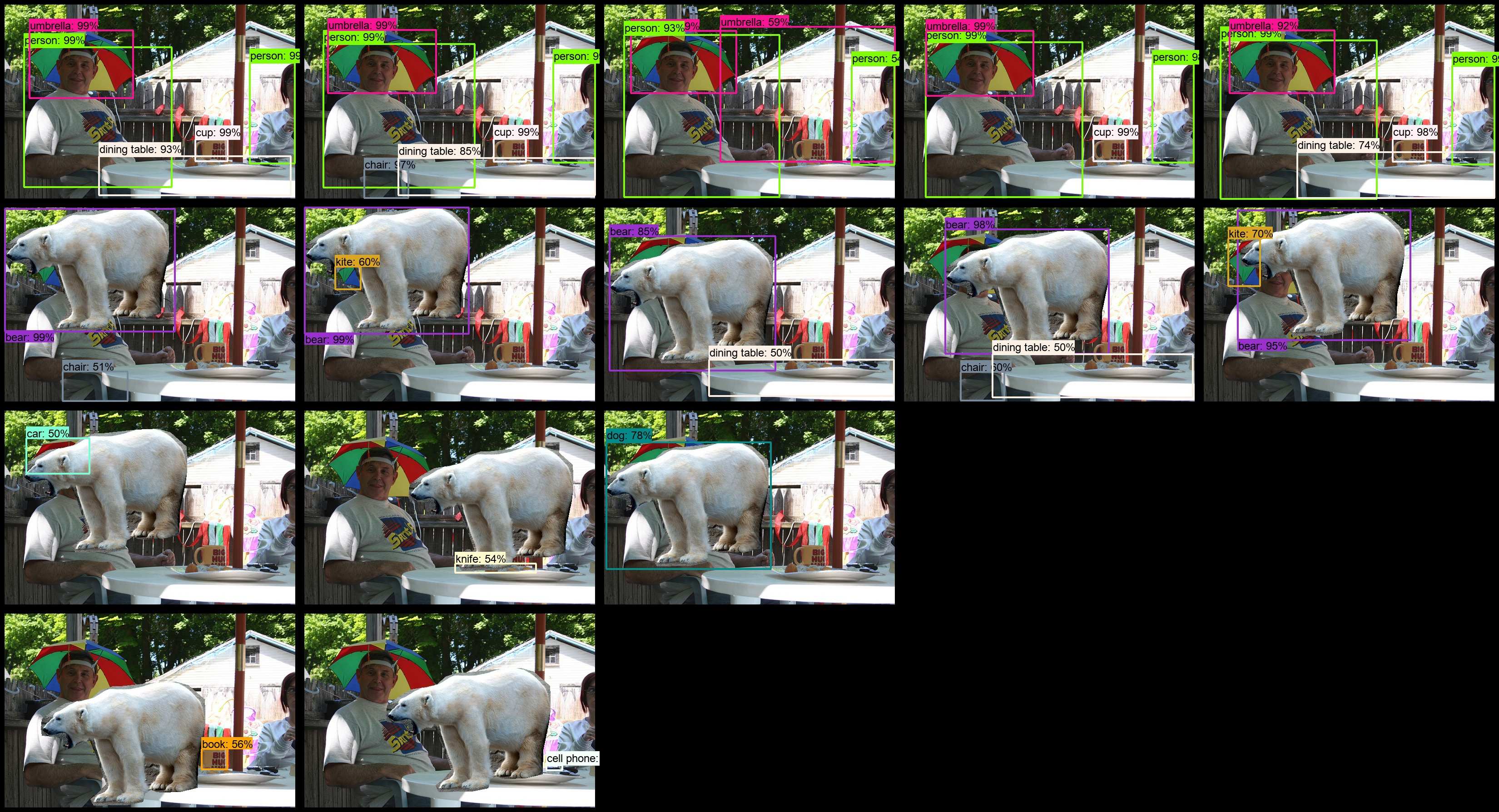}

\caption{\label{2} Detection with Transplanted Objects. Top row : original
images. Left-to-right: detection of models: faster\_rcnn\_inception\_resnet\_v2\_atrous\_coco,
faster\_rcnn\_nas\_coco, ssd\_mobilenet\_v1\_coco, mask\_rcnn\_inception\_resnet\_v2\_atrous\_coco,
mask\_rcnn\_resnet101\_atrous\_coco. Each row shows only newly added
detection w.r.t the previous row in the same column to avoid clutter.
Transplanting the bear causes a variety of new objects to be detected,
e.g.: chair, car, book (first column); kite, knife, cellphone (second
column).}
\end{figure*}

\begin{figure*}
\includegraphics[width=1\textwidth]{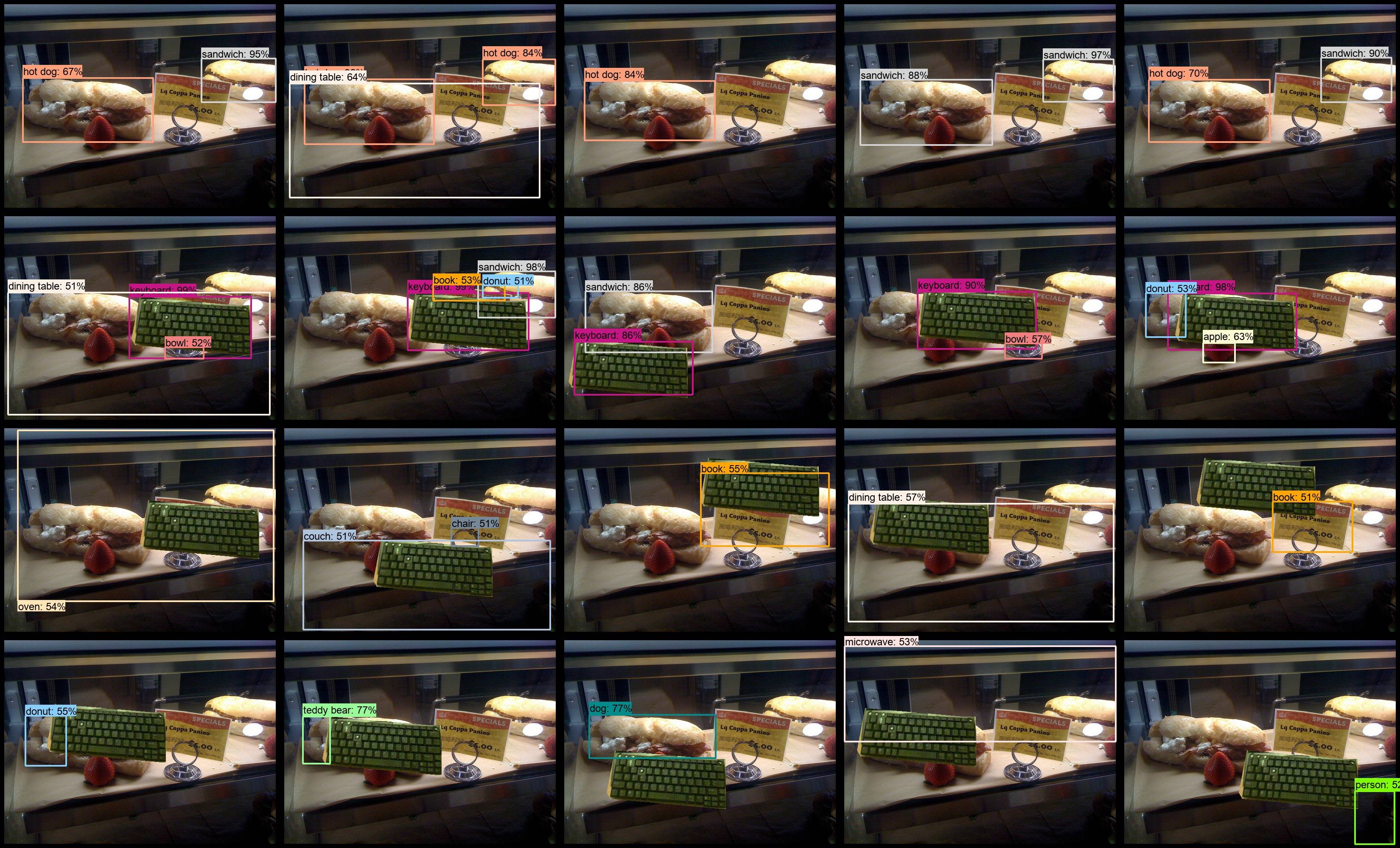}

\caption{\label{3}Detection with Transplanted Objects. Top row : original
images. Left-to-right: detection of models: faster\_rcnn\_inception\_resnet\_v2\_atrous\_coco,
faster\_rcnn\_nas\_coco, ssd\_mobilenet\_v1\_coco, mask\_rcnn\_inception\_resnet\_v2\_atrous\_coco,
mask\_rcnn\_resnet101\_atrous\_coco. Each row shows only newly added
detection w.r.t the previous row in the same column to avoid clutter.}
\end{figure*}

\begin{figure*}
\includegraphics[width=1\textwidth]{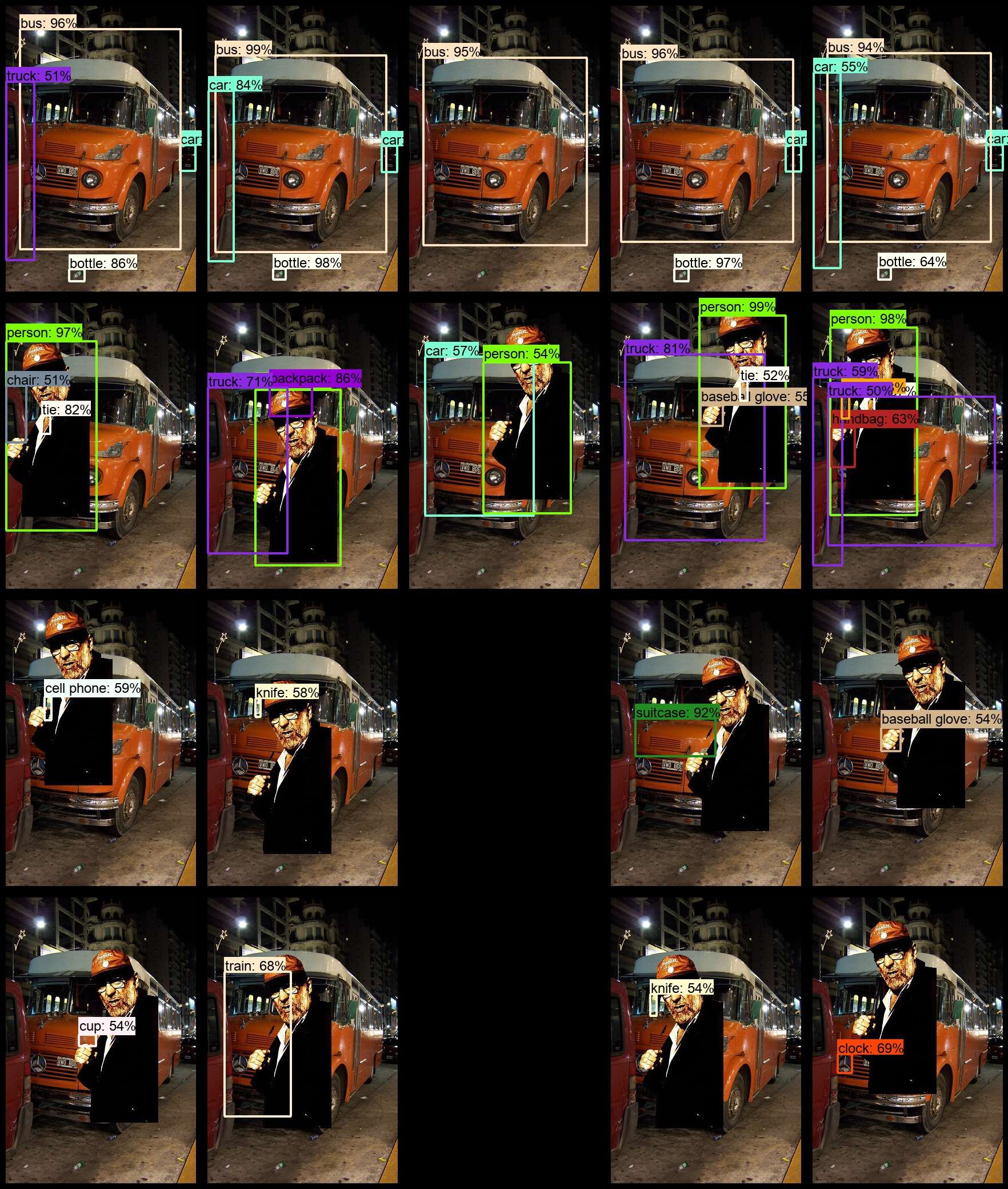}

\caption{\label{4}Detection with Transplanted Objects. Top row : original
images. Left-to-right: detection of models: faster\_rcnn\_inception\_resnet\_v2\_atrous\_coco,
faster\_rcnn\_nas\_coco, ssd\_mobilenet\_v1\_coco, mask\_rcnn\_inception\_resnet\_v2\_atrous\_coco,
mask\_rcnn\_resnet101\_atrous\_coco. Each row shows only newly added
detection w.r.t the previous row in the same column to avoid clutter.}
\end{figure*}

\begin{figure*}
\includegraphics[width=1\textwidth]{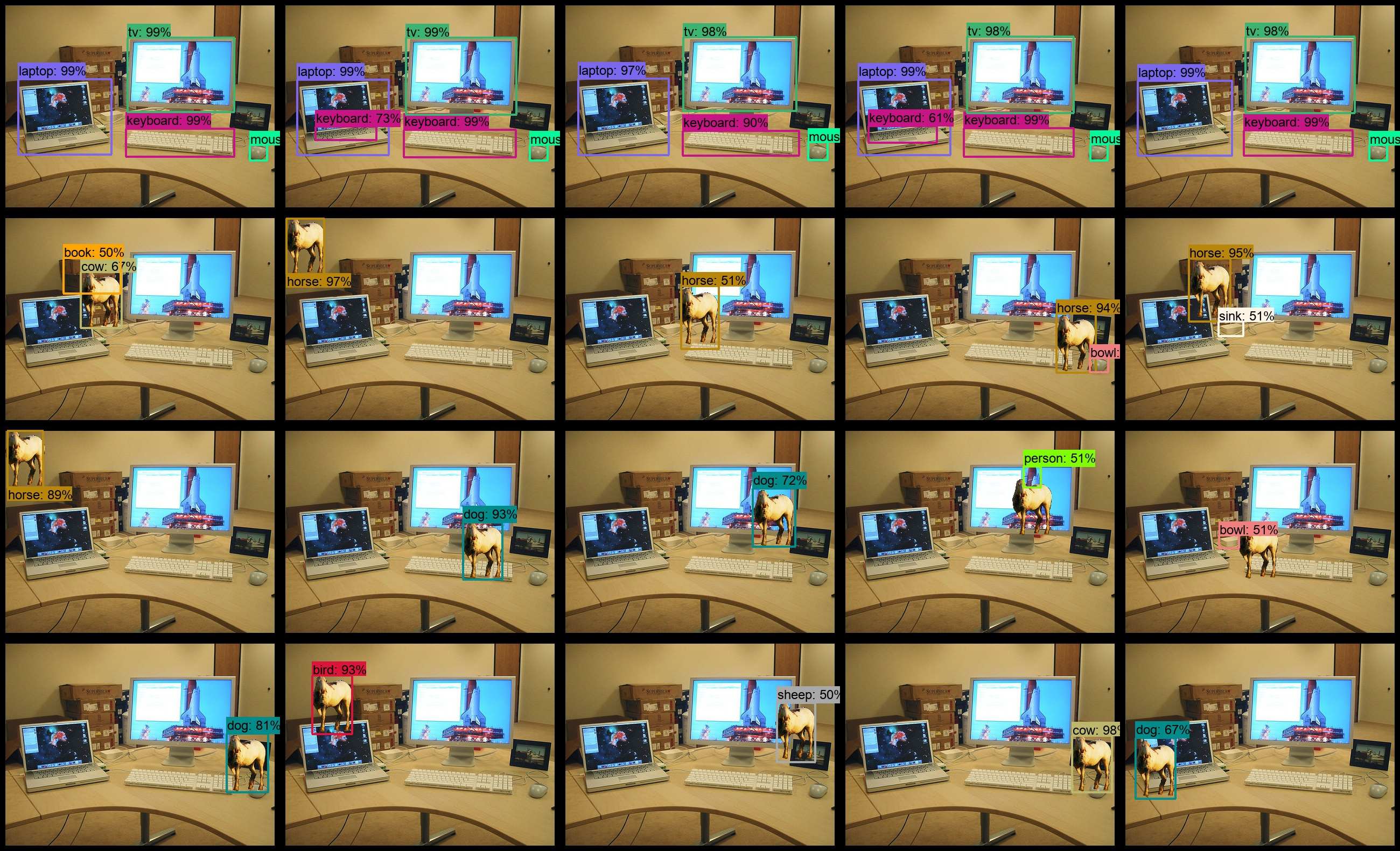}

\caption{\label{5}Detection with Transplanted Objects. Top row : original
images. Left-to-right: detection of models: faster\_rcnn\_inception\_resnet\_v2\_atrous\_coco,
faster\_rcnn\_nas\_coco, ssd\_mobilenet\_v1\_coco, mask\_rcnn\_inception\_resnet\_v2\_atrous\_coco,
mask\_rcnn\_resnet101\_atrous\_coco. Each row shows only newly added
detection w.r.t the previous row in the same column to avoid clutter.}
\end{figure*}

\begin{figure*}
\includegraphics[width=1\textwidth]{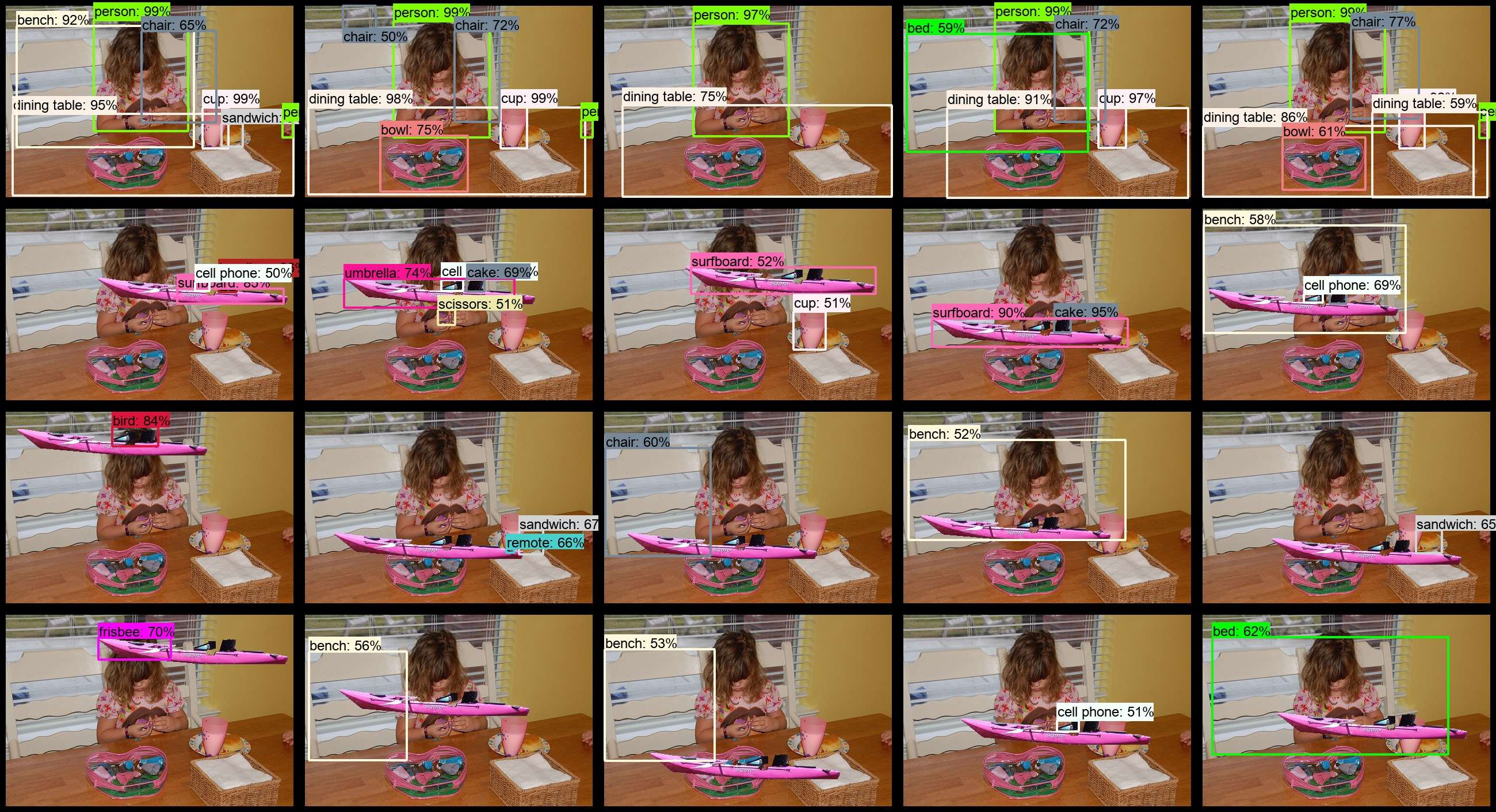}

\caption{\label{6}Detection with Transplanted Objects. Top row : original
images. Left-to-right: detection of models: faster\_rcnn\_inception\_resnet\_v2\_atrous\_coco,
faster\_rcnn\_nas\_coco, ssd\_mobilenet\_v1\_coco, mask\_rcnn\_inception\_resnet\_v2\_atrous\_coco,
mask\_rcnn\_resnet101\_atrous\_coco. Each row shows only newly added
detection w.r.t the previous row in the same column to avoid clutter.}
\end{figure*}

Figures \ref{2},\ref{3},\ref{4},\ref{5},\ref{6} show detailed
results of ranking a handful of the selected images according to the
``Class-Matching'' Criteria. We recommend viewing these figures
online with zooming, as they contain many notable details. The first
row of each of the figures shows the result on an unmodified image
by the detectors. For each column, each successive row shows a modified
image with a transplanted object, along with detection of objects
whose class did not appear among the detected classes in the previous
images. The order of the detectors, from left to right, is specified
in the figure caption and follows that of Table \ref{tab:Models-used-in}. 

Let us examine Figure \ref{3} as an interesting case. We pick to
address the effects in the strongest detector, \textbf{faster\_rcnn\_nas\_coco},
whose mAP is reported as 43\% on the MS-COCO dataset. This is a relatively
``heavy'' detector, requiring 1.83s on a Titan-X GPU. For comparison,
the second-best model, \textbf{faster\_rcnn\_inception\_resnet\_v2\_atrous\_coco
}requires 600ms per image, roughly a third of the time, with an AP
of 37\%. The results of \textbf{faster\_rcnn\_nas\_coco }are shown
in the second column. We only show detection results with a confidence
value that exceeds 0.5. 

The original detection (top row, second column) shows a couple of
detected hot-dogs, with the table detected as well.

The second row shows the result of adding a keyboard at a certain
location. The keyboard is detected with high confidence, though now
one of the hot-dogs, partially occluded, is detected as a sandwich
and a doughnut. What remains visible of the small sign is now detected
as a book. 

The third row shows that the region below the table is now interpreted
as a couch and the partially-occluded sign-holder is detected as a
chair. In the last row the left hot-dog is interpreted as a teddy-bear. 

We can see similar trends for the same detector in other images: in
Figure \ref{2}, the bear changes the interpretation of the image
so that a new kite (2nd row), knife (3rd row), cellphone (4th row)
are detected for different configurations. This also happens with
detectors of lower average performance. We deliberately chose to exemplify
this for what is likely one of the strongest detectors currently available,
showing results on additional ones for reference.

\subsection*{Co-occurring Objects}

So far, we have shown results where the pair of images and object
to be transplanted are selected randomly. Arguably, it is too much
to expect a network which has never seen a certain combination of
two categories within the same image to be able to successfully cope
with such an image at test time. We do not believe that requiring
each pair of object categories to co-occur in the training set is
a reasonable one, both practically and theoretically. Certainly, it
is not too much for a human. Out of context, humans are able to recognize
objects, though it requires more time \cite{biederman1972perceiving}. 

\begin{figure*}
\begin{centering}
\subfloat[]{\begin{centering}
\includegraphics[height=0.6\textheight]{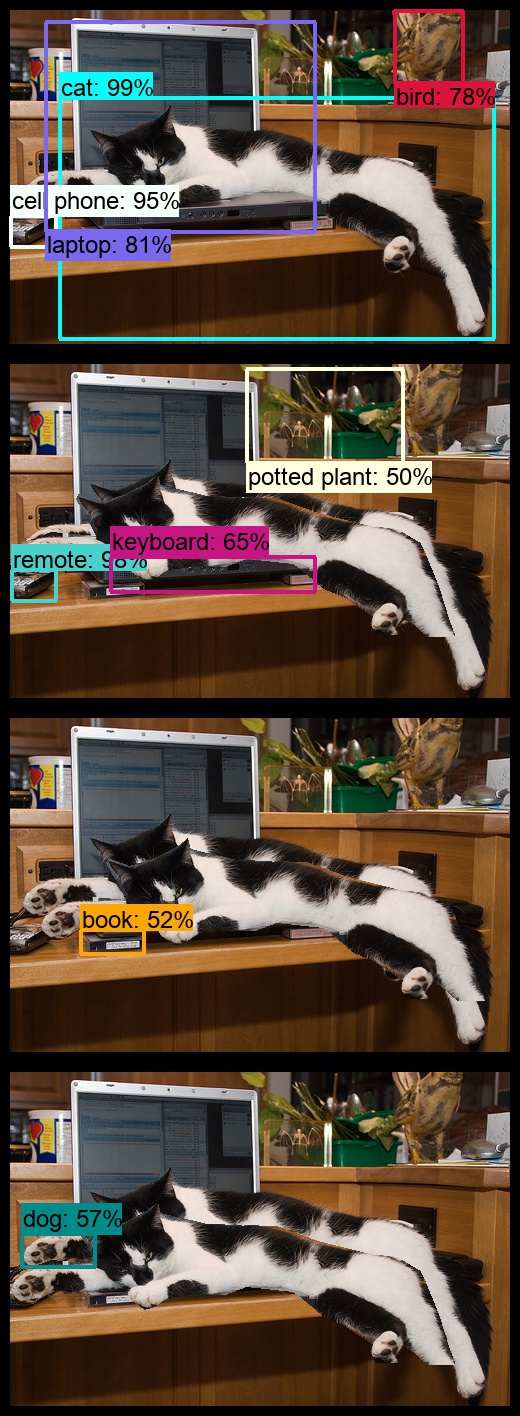}
\par\end{centering}

}\subfloat[]{\begin{centering}
\includegraphics[height=0.6\textheight]{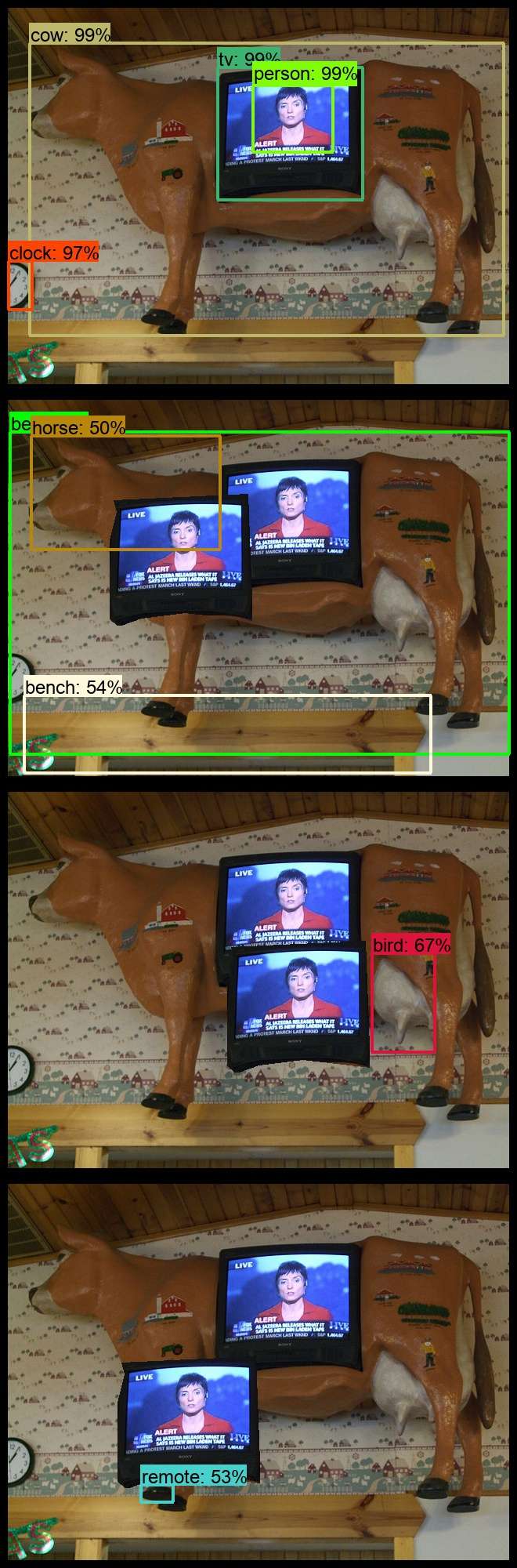}
\par\end{centering}

}\subfloat[]{\begin{centering}
\includegraphics[height=0.6\textheight]{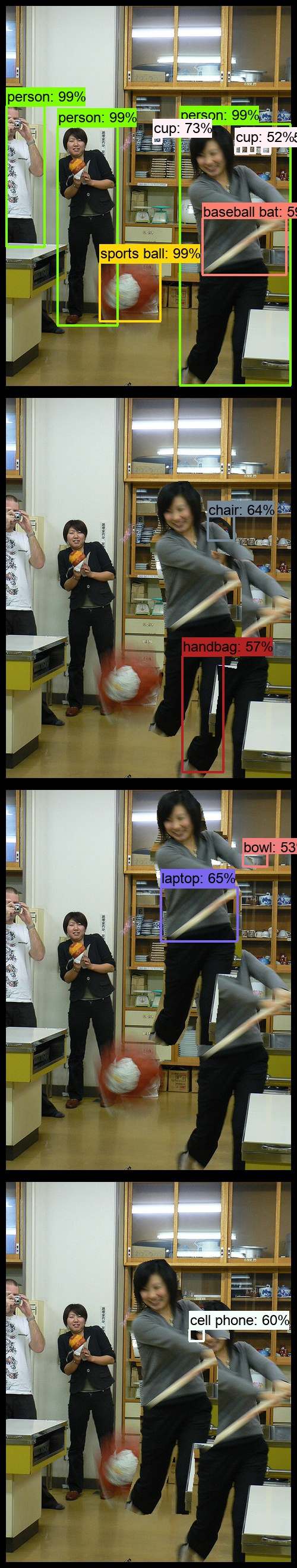}
\par\end{centering}

}\subfloat[]{\begin{centering}
\includegraphics[height=0.6\textheight]{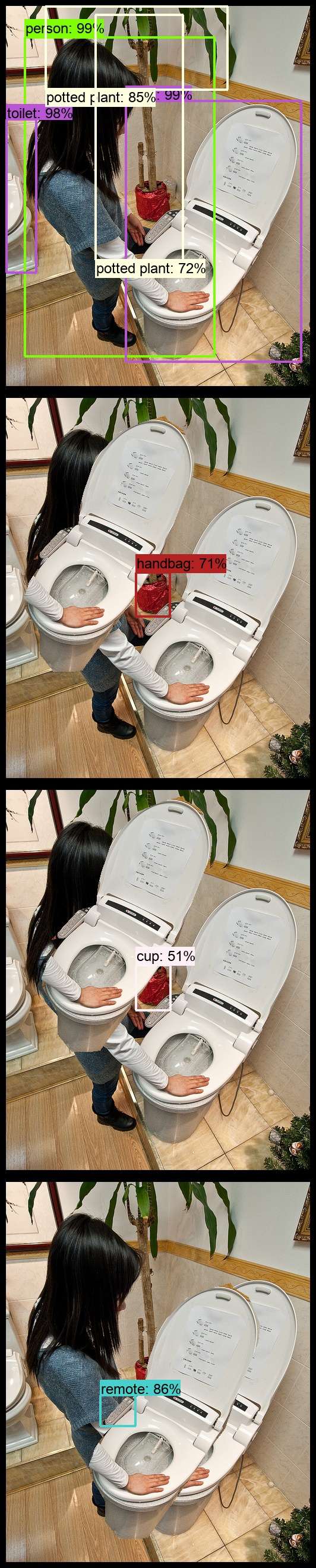}
\par\end{centering}

}\caption{\label{fig:duplicate}Effects of transplanting an object from an image
into another location in the same image. Top row: original detection.
Each subsequent rows: newly detected objected w.r.t to previous row,
induced by the translated object copy. }
\par\end{centering}
\end{figure*}

Nevertheless, we now turn to generating images of another extreme:
we duplicate an object from within an image and copy it to another
location \emph{in the same image. }Figure \ref{fig:duplicate} shows
some results of detection for generated images for 4 randomly picked
images. We see that the effect also happens for such images. Partial
occlusions and context seem to play a role here. For example, in column
(b), bottom row, the foot of the cow becomes a ``remote'' when near
a TV. The bottom if the plant (column d, last rows 2-3) are interpreted
as either a hand-bag or a cup, when part of the plant is occluded,
but a person's hand is nearby. The results in Figure \ref{fig:duplicate}
were all generated using the \textbf{faster\_rcnn\_nas\_coco }model. 

\subsection*{Feature Interference}

\begin{figure*}
\begin{centering}
\subfloat[]{\includegraphics[width=0.24\textwidth]{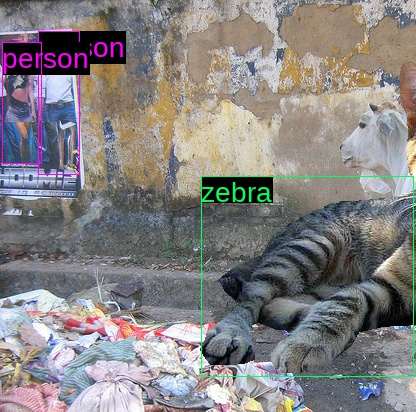}}\,\subfloat[]{\includegraphics[width=0.24\textwidth]{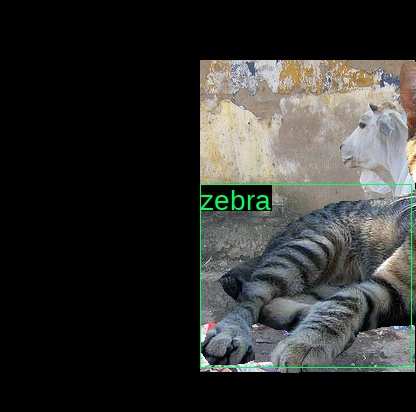}}\,\subfloat[]{\includegraphics[width=0.24\textwidth]{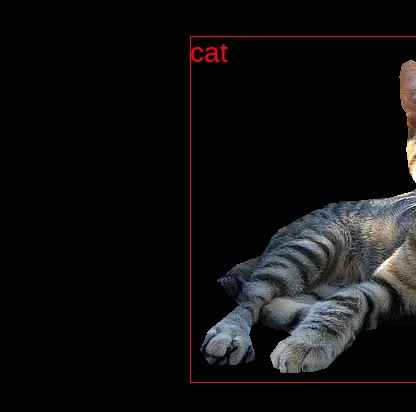}}\,\subfloat[]{\includegraphics[width=0.24\textwidth]{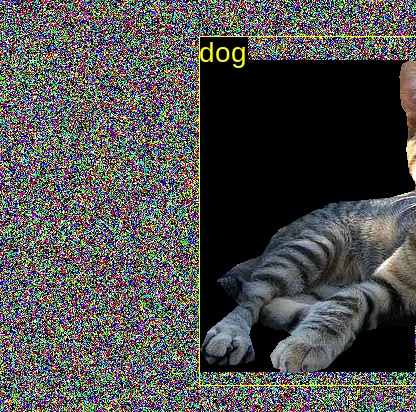}}
\par\end{centering}
\caption{\label{fig:entanglement}Feature Interference. A partially visible
cat is detected as a zebra (\emph{a}). Discarding all pixels \emph{outside
}the detection's bounding box does not fix the object's classification,
showing that features inside the region-of-interest (ROI) can cause
confusion (\emph{b}). Discarding also all non-cat pixels inside the
ROI leads to a fixed classification (\emph{c}). Adding random noise
in the range outside the bounding box once again makes the detection
incorrect , showing the effect of features outside the ROI (\emph{d})}
\end{figure*}

We now show how feature interference could be harmful to the detection
process, as a plausible explanation to the observed detection error.
As an example, consider the detection result in Figure \ref{fig:entanglement}
(a). A partially visible cat is detected and classified as a zebra.
We demonstrate that features attained from pixels that do not belong
to the actual object (cat) have an effect on the assigned class. This
is true both for pixels inside the ROI (region-of-interest) of the
object and for those outside of it: in Figure \ref{fig:entanglement}
(b) we set to zero all of the pixels outside of the bounding box.
The detection result is not changed. When we also zero out the pixels
inside the bounding box, leave those that belong to the cat, the resulting
label becomes ``cat''. This shows the effect of the pixels inside
the ROI. However, when we randomize the background intensity outside
the ROI, the label becomes ``dog''. This shows that features from
outside the ROI affect the final result of the detection. This experiment
was performed with a PyTorch port of the method of Yolov3 \cite{redmon2018yolov3},
which is very fast and yields results which are on par with state-of-the-art
in object detection. The final classification in this case relies
on features from a single grid-cell of a convolutional layer. 

\subsection*{Further Statistics}

To gain some more insight about the spread of the reported phenomena,
we take a few images and calculate statistics to summarize what happens
as the transplanted object is translated over a dense set of locations,
with a 5 pixels stride for each. We do this for 29 different images. 

First, we count how many times the interpretation of the scene has
changed from what is expected. This is done by matching the set of
bounding boxes produced by the detector in the original and modified
images. Recall that the set of detections of a modified image $I_{t}^{x,y}$
is denoted by $\mathcal{D}_{x,y}$ and those of the original image
$I$ by $\mathcal{D_{\emptyset}}$. 

We seek a maximal match between the bounding boxes of $\mathcal{D}_{x,y}$
and those of $\mathcal{D_{\emptyset}}$. A bipartite graph $\mathcal{G}$
is defined over the bounding boxes of both detection sets as the nodes.
Each node corresponding to a box from $\mathcal{D}_{x,y}$ has an
edge to a node from $\mathcal{D_{\emptyset}}$ only if their classes
are identical. The weight $w$ is the overlap score of the corresponding
boxes. A maximally weighted match $M_{x,y}$ is found. The score of
the match is calculated by 
\begin{equation}
S(M_{x,y})=\frac{\sum_{w\in M_{x,y}}w}{max(\left|\mathcal{D}_{x,y}\right|-1,\left|D_{\emptyset}\right|)}
\end{equation}

Where $\left|\cdot\right|$ is the cardinality of a set. This enforces
good matches both in set size and location of bounding boxes. The
highest attainable score is one. The score $S$ gives us the average
overlap score between each box and its matched one if there is a perfect
pairing between boxes and a lower score if some boxes are missing
in either set. 

This allows us to count the number of objects whose detection was
not affected by the transplanted object $T$. When counting $\left|\mathcal{D}_{x,y}\right|$
we in fact reduce 1 so the score will not be reduced by the detection
of $T.$

For each image with a translation $I_{t}^{x,y}$, we count the image
as ``affected'' if the $T$ at the corresponding translation gave
rise to a match score below a certain threshold $\tau$. The average
number of affected translations is reported in Table \ref{tab:Average-effect-of}
for varying values of the threshold $\tau$. A higher value of $\tau$
is a more strict matching criteria. In this context, an threshold
of 0.5 is quite a loose one, as we are trying to match two detections
of the exact same object in the original and modified image. For very
strict thresholds, i.e, $\tau=0.99$ we see that there is a very low
number of bounding boxes matched the original ones exactly (less than
25\%). The second row of Table \ref{tab:Average-effect-of} (\textbf{Affected-class-Agnostic})
also shows the number of affected locations if we allow two bounding
boxes to match even if their classes do not. By construction this
creates fewer mismatches, however not by a large margin. 

\begin{table}
\bigskip{}

\begin{centering}
\resizebox{\columnwidth}{!}{%
\begin{tabular}{cccccc}
\toprule 
 & $\tau=.3$ & $\tau=.5$ & $\tau=.7$ & $\tau=.95$ & $\tau=.99$\tabularnewline
\midrule
\midrule 
\%Affected & 10.3 & 20.6 & 31.3 & 53.7 & 75.6\tabularnewline
\midrule
\midrule 
\%Affected-class-Agnostic & 6.6 & 15.9 & 25.7 & 52.6 & 75.5\tabularnewline
\midrule
\midrule 
\%Affected-Occ-20 & 3.1 & 6.7 & 8.2 & 22.6 & 51.3\tabularnewline
\midrule
\midrule 
\%Affected-No-Occ & 2.4 & 4.5 & 4.95 & 9 & 22.4\tabularnewline
\bottomrule
\end{tabular}}
\par\end{centering}
\caption{\label{tab:Average-effect-of}Average effect of transplanting objects.
We show the average percentage of locations where the transplanted
object has caused the detection of any of the original objects to
be modified, with varying strictness of matching original and modified
detections (\textbf{Affected}). The threshold $\tau$ is the minimal
overlap to count two bounding boxes of the same class as a match.
A higher $\tau$ is a more strict matching criterion, leading to less
matches (more affected locations). For a majority of the translations,
there is no exact match between the detections on the modified and
original images. \textbf{Affected-class-Agnostic}: results for class-agnostic
matching between the bounding boxes. \textbf{Affected-Occluded-20}:
results where we count only cases where at most 20\% of the area of
each original object was covered by $T$. \textbf{Affected-No-Occ}:
results where $T$ occludes no object whatsoever.}
\end{table}

\subsubsection*{Occlusions}

The previous analysis did not consider that the object $T$ may occlude
partially or fully, many of the objects in the image. Therefore, we
calculate again the matching between the sets of bounding boxes while
recording to which extent each of the original objects was covered
by $T$. For each original bounding box $b\in D_{\emptyset}$, we
calculate the coverage of \textbf{$b$} by $T$ as

\begin{equation}
C_{b,T}=\frac{|b\cap T|}{|b|}
\end{equation}

Where $|\cdot$\textbar{} corresponds to the area of the bounding
box (in this context $T$ is interpreted as $T$'s bounding box).
We calculate the maximal coverage 
\begin{equation}
C_{T}=\max_{b\in D_{\emptyset}}C_{b,T}
\end{equation}

We calculate again the number of affected objects, this time discarding
each image for which the maximal coverage $C_{T}$ exceeds 0.2. In
other words, we discard all images where the object $T$ covered any
object's area by more than 20\%. The results are displayed in the
third row of Table \ref{tab:Average-effect-of} (\textbf{Affected-Occ-20}).
The last row (\textbf{Affected-No-Occ}) shows the results of discarding
all of the images where $T$ did not cover any of the original detections.
Even in these two cases, where the objects are hardly touched by $T$
- or not at all - the object transplant still has a non-negligible
effects.

\subsection*{Global Effects Beyond Detection}

\begin{figure}
\subfloat[]{\includegraphics[width=1\columnwidth]{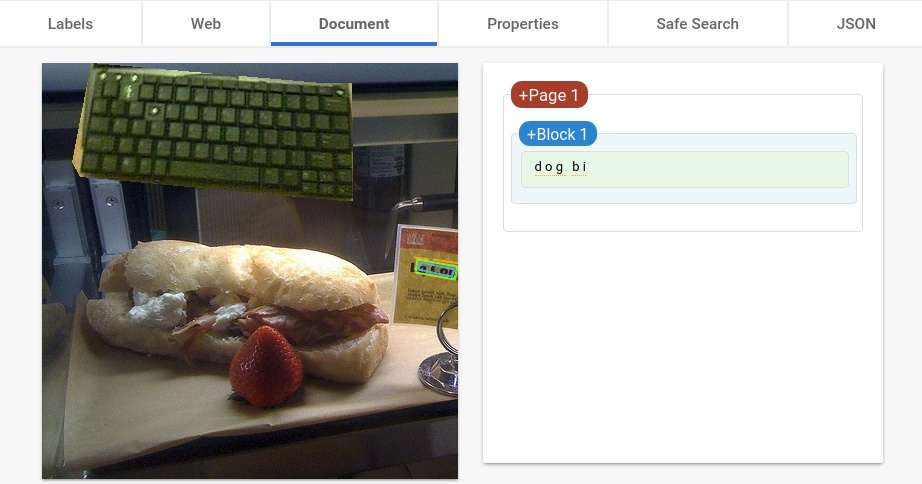}

}

\subfloat[]{\includegraphics[width=1\columnwidth]{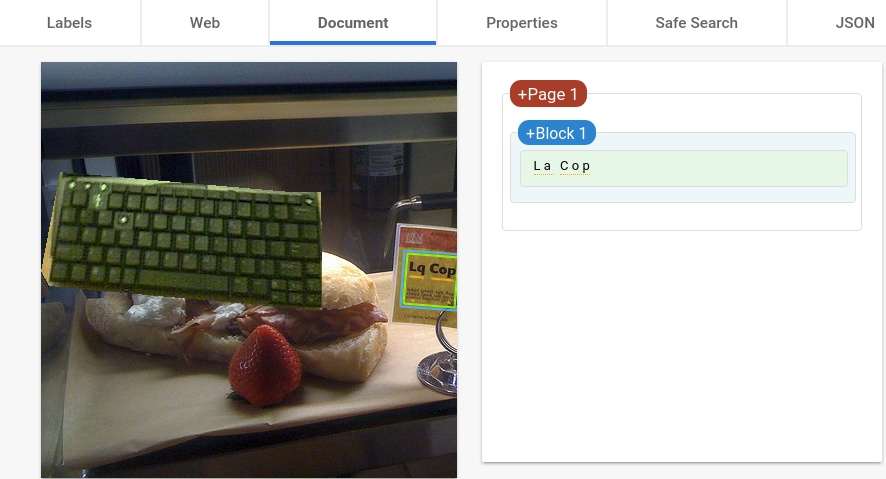}

}

\caption{\label{fig:ocr}Non-local effects of object transplant on Google's
OCR. A keyboard placed in two different locations in an image causes
a different interpretation of the text in the sign on the right. The
output for the top image is ``dog bi'' and for the bottom it is
``La Cop''}
\end{figure}

In a preliminary experiment, we uploaded a couple of the images where
no object was detected at all to Google's Vision API website\footnote{\url{https://cloud.google.com/vision/}}.
These image were picked arbitrarily. We report the result here as
we find it noteworthy for further exploration. It seems that the OCR
portion of their method also exhibits a surprising amount of non-local
sensitivity to transplanted objects. Figure \ref{fig:ocr} shows this:
the keyboard is placed in two different locations in the image. Though
each of the locations is such that the keyboard is far from the sign,
the interpretation of the sign is different in each case. 

\section*{Discussion}

We now raise several possible reasons for the observed behaviour of
current object-detectors. Though there are several reported phenomenon
here, we believe that there they are not independent and that some
(but not all) share common underlying reasons. 

\subsubsection*{Partial Occlusions}

It is quite widely accepted that partial occlusions were and still
are a challenge to object detectors. A good sign of generalization
is being able to cope with partial occlusions. Indeed, in many of
the examples that we tested, the modern object detectors were quite
robust to such occlusions. In \cite{wang2017fast}, this is acknowledged
and a data-driven solution is proposed, implemented via an adversarial
network to generate examples including occlusions and deformations.
Recently, Zhang et al. \cite{zhang2018deepvoting} proposed a method
to vote using local evidence in order to localize semantic parts even
in heavily occluded images. 

\subsubsection*{Out of Distribution Examples}

It is possible that the modifies images have very low likelihood to
occur under the distribution of images under the training set. Since
we ``paste'' objects using their foreground mask onto the target
image, abrupt edges are created at the patches of the object's border.
These edges may be out-of-distribution when considering the local
appearance of naturally occurring edges. 

The images generated here could be viewed as a variant of adversarial
examples \cite{szegedy2013intriguing}, in which small image perturbations
(imperceptible to humans) cause a large shift in the network's output.
The images we generate are of a somewhat opposite flavor: while we
do not limit the magnitude of the difference between the original
and modified image, the detectors are sometimes ``blind'' to the
inserted object, as demonstrated in Figure \ref{fig:Detecting-an-elephant}
(b,d,e,g,i). In addition, our examples are not ``targeted'' in the
sense that no optimization process is required to generate them; they
seem prevalent enough so that a simple scan of transplanting translated
versions of one object in the other can give rise to multiple wrong
interpretations.

\subsubsection*{(Lack of) Signal Preservation}

Spatial pooling, prevalent in deep neural networks, is useful for
reasons of efficiency and invariance to certain spatial deformations.
However, as recently observed by Azulay and Weiss \cite{azulay2018deep},
these pooling layers actually prevent the network from being truly
shift-invariant. Such behaviour is in fact anticipated by simple signal-processing
considerations and has in fact been discussed much earlier \cite{simoncelli1992shiftable}.
The authors of \cite{azulay2018deep} also observe that small image
shifts, as well as other geometric transforms as scaling, can cause
the network's output to change dramatically. 

\subsubsection*{Contextual Reasoning}

It is not common for current object detectors to explicitly take into
account context on a semantic level, meaning that interplay between
object categories and their relative spatial layout (or possibly additional)
relations) are encoded in the reasoning process of the network. Though
many methods claim to incorporate contextual reasoning, this is done
more in a feature-wise level, meaning that global image information
is encoded somehow in each decision. This is in contrast to older
works, in which explicit contextual reasoning was quite popular (see
\cite{choi2012context} for mention of many such works). Still, it
is apparent that some implicit form of contextual reasoning does seem
to take place. One such example is a person detected near the keyboard
(Figure \ref{3}, last column, last row). Some of the created images
contain pairs of objects that may never appear together in the same
image in the training set, or otherwise give rise to scenes with unlikely
configurations. For example, non co-occurring categories, such as
elephants and books, or unlikely spatial / functional relations such
as a large person (in terms of image area) above a small bus. Such
scenes could cause misinterpretation due to contextual reasoning,
whether it is learned explicitly or not. 

\subsubsection*{Non-Maximal Suppression}

Many changes to the detection results seem to be non-local. Whereas
partial occlusions can be regarded as local effects of the transplanted
object that directly change the object's appearance, we also see sometimes
changes in detection of objects that are far away from the transplanted
object. We suggest that this may be partially due to the process of
non-maxima suppression (NMS) common in object detectors: assume that
a previously detected object $A$ is no longer detected due to a partial
occlusion by the transplanted object $T$. An object $B$, which overlaps
with $A$ and was previously suppressed during the NMS, would possibly
not be suppressed now. Similar chain-reactions could cause the NMS
process to eventually affect a far away object that is not adjacent
to $T$. 

\subsubsection*{Feature Interference}

Modern object detectors use features obtained from a convolutional
layer in order to generate the final object class and bounding box
prediction. These regions are either fixed in size \cite{liu2016ssd,redmon2018yolov3}
or rectangular. The ROI-Pooling operation \cite{girshick2015fast}
performs max-pooling features from sub-windows of a convolutional
feature map over a region of interest (ROI). This operation is affected
by the following facts:
\begin{enumerate}
\item The region of interest is a rectangular one. This means that sections
of the region that do not belong to the object are also pooled, including
background appearance as well as appearance of the object. 
\item Each part of the feature map can have a large effective receptive
field. In practice, this means that features are pooled from outside
the bounding box of the detected object.
\end{enumerate}
On one hand, including features from an object's surroundings could
provide useful contextual cues to improve object detection, especially
for objects that do not provide enough evidence due to size, partial
occlusion, etc. On the other hand, invariably mixing additional features
into the final classification score could hinder the result. 

We believe that feature interference, demonstrated in Figure \ref{fig:entanglement},
is perhaps the root cause for most of the observed phenomena, and
that effects that seem due to partial occlusion or contextual reasoning
are specific cases of this problem. Experiments discussing this difficulty
have been introduced in Rosenblatt \cite{rosenblatt1961principles}.
The associated problem in biological vision has later been coined
the ``binding problem'' \cite{von1999and} and noted in cognitive
studies in humans \cite{treisman1982illusory} as well. The works
of Tsotsos et al. refer to this issue as cross-talk \cite{tsotsos1995modeling,tsotsos2017complexity}
and predicts neural mechanisms that are in place to overcome this
issue, in the form of the Selective Tuning framework. In brief, the
idea is that once a first pass is finished through the visual hierarchy,
the dominant signal propagates downwards through the hierarchy, performing
spatial and feature attenuation so that the next pass of the signal
will contain information on the object of interest that is less entangled
with surrounding features. The model is discussed in more extensive
details in \cite{tsotsos2008different} , describing the different
stages of information flow through the visual hierarchy. We suggest
that such mechanisms are expected to alleviate many of the observed
phenomena, and leave this for future development. 

\paragraph*{Acknowledgment}

This material is based upon work supported by the Air Force Office
of Scientific Research under award number FA9550-18-1-0054, and by
the Canada Research Chairs program through grants to JKT for which
the authors are grateful. 

\noindent \begin{center}
\begin{table}
\begin{centering}
\begin{tabular}{cc}
\toprule 
Model Name & COCO mAP\tabularnewline
\midrule
\midrule 
faster\_rcnn\_inception\_resnet\_v2\_atrous\_coco & 37\tabularnewline
\midrule 
faster\_rcnn\_nas\_coco & \textbf{43}\tabularnewline
\midrule 
ssd\_mobilenet\_v1\_coco & 21\tabularnewline
\midrule 
mask\_rcnn\_inception\_resnet\_v2\_atrous\_coco & 36\tabularnewline
\midrule 
mask\_rcnn\_resnet101\_atrous\_coco & 33\tabularnewline
\bottomrule
\end{tabular}
\par\end{centering}
\caption{\label{tab:Models-used-in}Models used in the reported experiments,
along with their mean average precision. }
\end{table}
\par\end{center}

\bibliographystyle{ieee}
\bibliography{elephant}

\end{document}